\documentclass[journal]{IEEEtran}
\ifCLASSINFOpdf
\else
\fi
\usepackage[utf8]{inputenc}
\usepackage{wrapfig}
\usepackage{tabularx}
\usepackage{booktabs}

\usepackage{amsmath}
\usepackage{graphicx}
\usepackage[colorinlistoftodos]{todonotes}
\usepackage[colorlinks=true, allcolors=blue]{hyperref}
\usepackage[noabbrev,capitalize]{cleveref}
\usepackage{url}
\usepackage{cleveref}
\usepackage{fancyhdr}
\usepackage{mdframed}
\usepackage{xcolor}
\usepackage{subcaption}

\definecolor{color0}{HTML}{A6CEE3}
\definecolor{color1}{HTML}{1F78B4}
\definecolor{color2}{HTML}{B2DF8A}
\definecolor{color3}{HTML}{33A02C}

\usepackage{pifont}

\newcolumntype{Y}{>{\centering\arraybackslash}X}

\begin{document}
%
\title{Deep Learning for Classification\\ of Hyperspectral Data: A Comparative Review}
%
%
%

\author{Nicolas~Audebert, Bertrand~Le~Saux,~\IEEEmembership{Member,~IEEE}
    and    S{\'e}bastien~Lef{\`e}vre
\thanks{N. Audebert and B. {Le Saux} are with DTIS, ONERA, University Paris Saclay, F-91123 Palaiseau - France. E-mails:  \mbox{nicolas.audebert@onera.fr}, \mbox{bertrand.le\_saux@onera.fr}.}
\thanks{S. Lef{\`e}vre is with Univ. Bretagne Sud, UMR 6074, IRISA, F-56000 Vannes, France. E-mail: sebastien.lefevre@irisa.fr}
\thanks{Manuscript received April 30, 2018.}}
\maketitle


\begin{abstract}
In recent years, deep learning techniques revolutionized the way remote sensing data are processed. Classification of hyperspectral data is no exception to the rule, but has intrinsic specificities which make application of deep learning less straightforward than with other optical data. This article presents a state of the art of previous machine learning approaches, reviews the various deep learning approaches currently proposed for hyperspectral classification, and identifies the problems and difficulties which arise to implement deep neural networks for this task. In particular, the issues of spatial and spectral resolution, data volume, and transfer of models from multimedia images to hyperspectral data are addressed. Additionally, a comparative study of various families of network architectures is provided and a software toolbox is publicly released to allow experimenting with these methods.\footnote{\url{https://github.com/nshaud/DeepHyperX}} 
This article is intended for both data scientists with interest in hyperspectral data and remote sensing experts eager to apply deep learning techniques to their own dataset.
\end{abstract}


%
\IEEEpeerreviewmaketitle

\section{Introduction}

Thanks to recent advances in deep learning for image processing and pattern recognition, remote sensing data classification progressed tremendously in the last few years. In particular, standard optical imagery (Red-Green-Blue --RGB-- and Infra-Red --IR--) benefited from using deep convolutional neural networks (CNN) for tasks such as classification, object detection or semantic segmentation~\cite{audebert_semantic_2016,volpi_dense_2017,marmanis_semantic_2016}. This was made possible by the transfer of models developed in computer vision, which focuses mostly on images encoded on three channels. However, remote sensing relies often on multispectral imagery (coming from satellites such as Sentinel-2 or Landsat, or from airborne sensors) which allows to capture simultaneously the radiance at several wavelength bands. Hyperspectral imaging (HSI) data are a subset of multispectral data for which the wavelength resolution is fine, wavelength bands are contiguous and their range is particularly high. It makes possible a precise analysis of soils and materials~\cite{bioucas-dias-hyper-GRSM13, cubero-castan_physics-based_2015}.


Indeed, the high spectral resolution allows to characterize precisely the electromagnetic spectrum of an object. However, most hyperspectral sensors have a low spatial resolution so deep learning techniques designed for computer vision are not easily transferable to hyperspectral data cubes since the spectral dimension prevails over the spatial neighborhood in most cases. If compared with today's optical images, the volume of data is similar but the structure is completely different. Moreover, the low spatial resolution actually limits the number of samples available for training statistical models. This also makes more difficult the annotation process which is required in supervised learning, since objects smaller than the spatial resolution are mixed with their neighborhood. These two points are the main challenges to take up in order to use deep learning for hyperspectral image processing.

This article aims at bridging the gap between data scientists and hyperspectral remote sensing experts. Thus, it is more focused than previous reviews on deep learning~\cite{zhu-deep-in-RS-GRSM2017} while presenting hyperspectral peculiarities from a deep learning point of view, different from~\cite{ghamisi-advances-HSI-overview-GRSM2017}. We first summarize some principles of hyperspectral imaging and list some reference datasets available for public use. We then review some standard machine learning techniques used for classification before focusing on recent works with deep learning, where the comparison of existing networks is supported by experimental analysis. We finally conclude with an emphasis on the emerging research axes.

\section{Hyperspectral Imaging: Principles and Resources}

\subsection{Hyperspectral imaging principles in a nutshell}


Hyperspectral sensors measure the intensity of the radiant flux for a given surface and a given wavelength, i.e. a physical quantity in watts per squared meter steradian ($\mathrm{W}/(\mathrm{sr}.\mathrm{m}^2)$). Precisely, per each surface unit (which corresponds to a pixel of the image) the sensor captures light emitted and reflected by the object as a spectrum of several hundreds of channels, which defines a spectral response curve. 



In the context of Earth Observation, signals coming from the Earth surface are changed by atmospheric perturbations such as clouds, water vapor atmospheric aerosols, etc. So, for remote sensing of surfaces and land cover, the reflectance is preferably used, defined as the ratio between the emitted flux of the surface and the incidental flux. This ratio gives the reflecting effectiveness of a given object for each light wavelength band. Reflectance is an intrinsic property of the materials, independently of the environment, and thus is highly discriminative for classification purposes.

To compensate for the atmospheric perturbations, several atmospheric correction methods have been developed~\cite{deschamps_atmospheric_1980,rahman_smac:_1994,chavez_image-based_1996} in order to obtain measures able to characterize the land cover~\cite{gao_atmospheric_2009}. They take the intensity images as input and produce reflectance images, by getting rid of the light diffusion effects and radiative phenomena of the atmosphere. They are based on elaborate physics models and take into account acquisition parameters such as the sunlight levels or the local digital surface model to deal with multiple-reflection phenomena.

In practice, hyperspectral images are $(w,h,B)$ tensors, i.e. three-dimensional cubes with two spatial dimensions (width $w$ and height $h$) and a spectral one (with $B$ bands). Compared to volumetric data, e.g. seismic data cubes, this so-called hypercube is anisotropic: the three dimensions do not represent the same physical displacement. However, all values in the hypercube are expressed in the same unit, either light intensities or reflectances, which makes linear operations on a 3D subset of the cube mathematically and physically valid. This property will come in handy when dealing with convolutions and filtering operations on the hypercube.


\subsection{Reference datasets}
\label{sec:datasets}




Several public datasets captured using an hyperspectral sensor have been made available to the scientific community. They usually come along with annotations to evaluate the classification performances. Pavia, Indian Pines and Data Fusion Contest 2018 are presented in details in the following. Other standard ones include Salinas (captured with the AVIRIS sensor over the Salinas Valley, Cal., USA and composed of classes such as vegetable cultures, vineyeard and bare soils), Kennedy Space Center (also with AVIRIS over the Kennedy Space Center, with classes such as wetlands and various types of wild vegetation) and Botswana (captured with the Hyperion sensor from the EO'1 satellite, with 14 classes of vegetation and swamps over the Okavango delta). While characteristics of all are listed and compared in Table~\ref{tab:hyperx_datasets}, a few ones are now presented in details.

\subsubsection{Pavia}

\begin{wrapfigure}{l}{0.25\textwidth}
  \vspace{-1em}
  \begin{center}
    \includegraphics[width=0.25\textwidth]{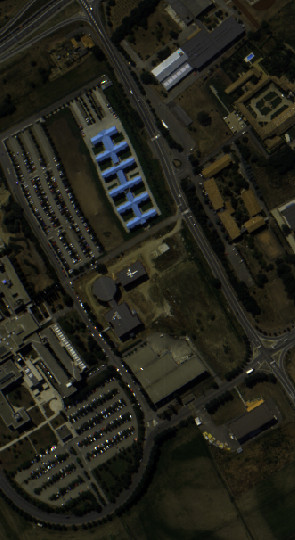}
  \end{center}
  \vspace{-1em}
  \caption{Pavia University (natural composite image). \label{fig:paviaU}}
  \vspace{-1em}
\end{wrapfigure}

Pavia is a dataset captured using the ROSIS sensor with a ground sample distance (GSD) of 1.3 m over the city of Pavia, Italy (cf. Fig.~\ref{fig:paviaU}). It is divided in two parts: Pavia University (103 bands, $610\times340$px) and Pavia Center (102 bands, $1096\times715$px). 9 classes of interest are annotated covering 50\% of the whole surface. They comprise various urban materials (such as bricks, asphalt, metals), water and vegetation.

This has been for long one of the main reference datasets because it is one of the largest labeled HSI data and because it allows to evaluate the use of HSI for potential applications. However, some preprocessing might be necessary in order to remove some pixels with no spectral information and a few errors occur in the ground-truth.







\subsubsection{Data Fusion Contest 2018 (DFC2018)}

\begin{figure}[!tbp]
\begin{center}
\includegraphics[width=0.45\textwidth]{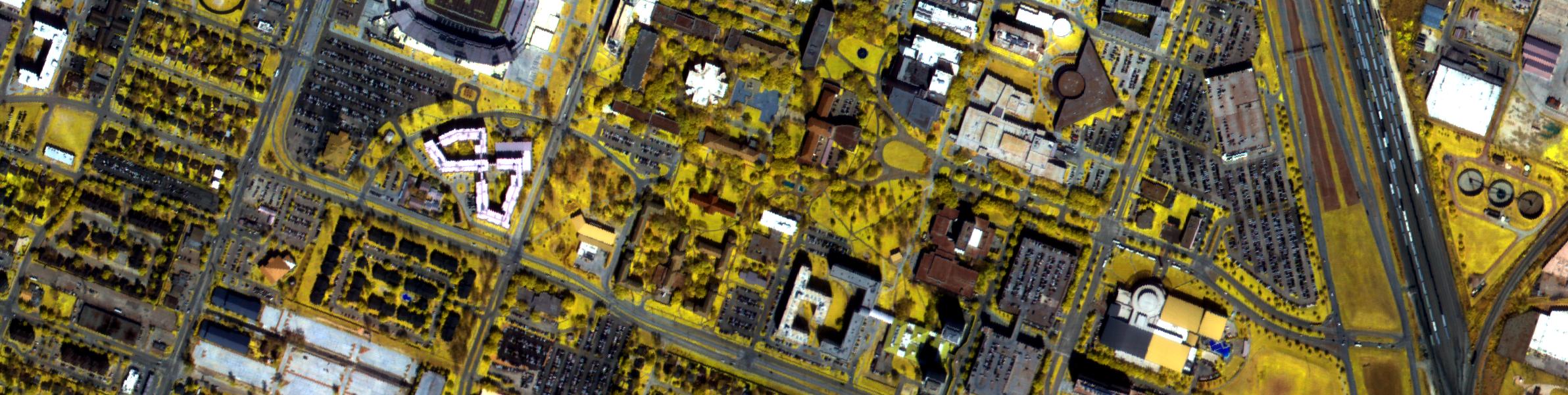} \\
\vspace{0.5em}
\includegraphics[width=0.45\textwidth]{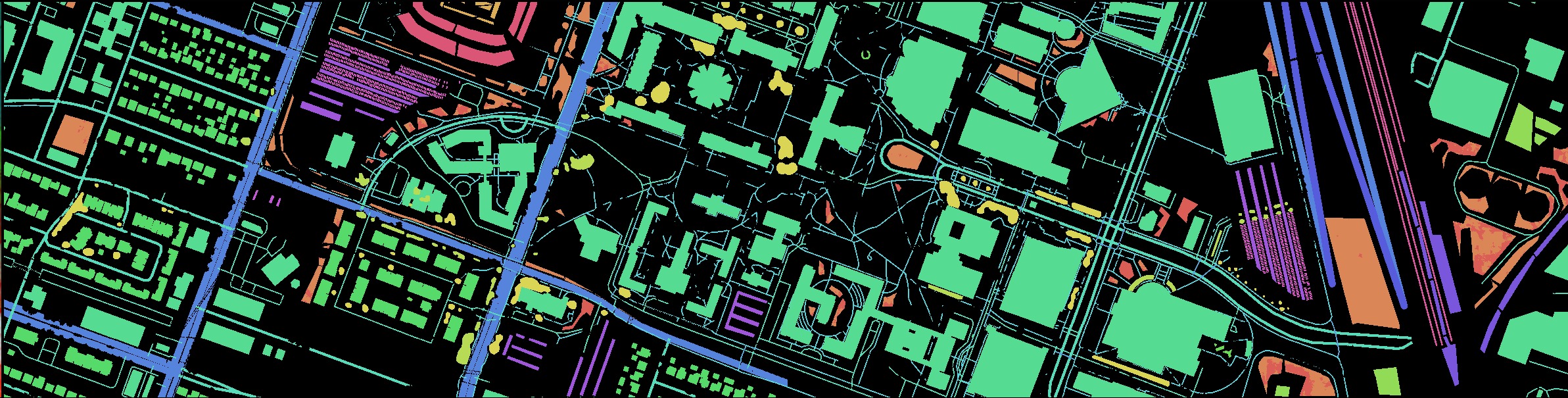}
\caption{Data Fusion Contest 2018~\cite{dfc2018} over Houston: composite image with bands 48, 32 and 16 (top) and ground-truth (bottom row).}
\label{fig:dfc2018}
\end{center}
\end{figure}

The DFC2018 hyperspectral data (cf. Fig.~\ref{fig:dfc2018}) was acquired over Central Houston, Texas, USA, using an airborne sensor. It covers a [380--1,050]~nm spectral range over 48 contiguous bands at 1~m GSD. 20 classes of interest are defined and include not only urban categories (buildings and roads of various types, railways, cars, trains, etc.) but also various vegetation types (stressed, healthy, deciduous or evergreen trees) and specific materials. This dataset is part of the 2018 Data Fusion Contest release, along with very-high-resolution imagery and multispectral LiDAR \cite{dfc2018,dfc2018-grsm}.

\subsubsection{Indian Pines}

\begin{wrapfigure}{r}{0.20\textwidth}
  \begin{center}
    \includegraphics[width=0.20\textwidth]{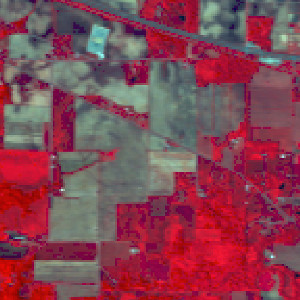}
  \end{center}
  \caption{Indian Pines (natural composite image).}
  \label{fig:indianpines}
\end{wrapfigure}

\begin{figure}
	\begin{subfigure}{0.49\textwidth}
      \begin{subfigure}{0.48\textwidth}
      \includegraphics[width=\textwidth]{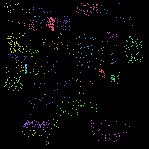}
      \caption*{Train (random)}
      \end{subfigure}
      \begin{subfigure}{0.48\textwidth}
      \includegraphics[width=\textwidth]{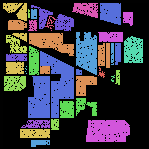}
      \caption*{Test (random)}
      \end{subfigure}
    \end{subfigure}
	\begin{subfigure}{0.49\textwidth}
    \begin{subfigure}{0.48\textwidth}
      \includegraphics[width=\textwidth]{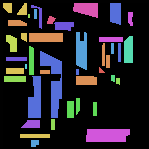}
      \caption*{Train (disjoint)}
      \end{subfigure}
      \begin{subfigure}{0.48\textwidth}
      \includegraphics[width=\textwidth]{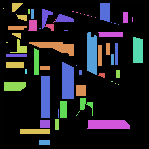}
      \caption*{Test (disjoint)}
      \end{subfigure}
    \end{subfigure}
    \caption{Examples of train/test splits on the Indian Pines dataset.}
    \label{fig:indianpines_gt}
\end{figure}

Indian Pines is a dataset captured using the AVIRIS sensor. The scene covers agricultural areas in North-Western Indiana, USA, with a ground sampling distance (GSD) of 20~m/px, resulting in a $145\times145$px image with 224 spectral bands. Most parts of the image represent fields with various crops while the rest denotes forests and dense vegetation. 16 classes (cf. Fig.~\ref{fig:indianpines} and~\ref{fig:indianpines_gt}) are labeled (e.g. corn, grass, soybean, woods, etc.), some of them being very rare (less than 100 samples for alfalfa or oats). Water absorption bands (104$\rightarrow$108, 150$\rightarrow$163 and 220) are usually removed before processing. In spite of its limited size, it is one of the main reference datasets of the community. Though, rare classes are usually not taken into account when evaluating classification algorithms.

\subsubsection{Dataset summary}

The various dataset statistics and informations are compiled in Table~\ref{tab:hyperx_datasets}.
This highlights that the main issue with applying machine learning approaches to hyperspectral data lies in the small number of available samples. Existing datasets are small with respect to standard optical imagery. Moreover, due to sensor diversity and post-processing methods, it is not possible to train algorithms simultaneously on different datasets.

\begin{table}[bht]
\caption{Main public labeled datasets in hyperspectral imaging.}
\label{tab:hyperx_datasets}\setlength{\tabcolsep}{1pt}
\begin{tabularx}{0.5\textwidth}{ c r r c r r c c }
\toprule
Dataset & \multicolumn{1}{c}{Pixels} & Bands & Range & \multicolumn{1}{c}{GSD} & \multicolumn{1}{c}{Labels} & Classes & Mode\\
\midrule
Pavia (U \& C) & 991,040 & 103 & 0.43-0.85 $\mu$m & 1.3 m & 50,232 & 9 & Aerial\\
Indian Pines & 21,025 & 224 & 0.4-2.5 $\mu$m & 20 m & 10,249 & 16 & Aerial\\
Salinas & 111,104 & 227 & 0.4-2.5 $\mu$m & 3.7 m & 54,129 & 16 & Aerial\\
KSC & 314,368 & 176 & 0.4-2.5 $\mu$m & 18 m & 5,211 & 13 & Aerial\\
Botswana & 377,856 & 145 & 0.4-2.5 $\mu$m & 30 m & 3,248 & 14 & Satellite\\
DFC 2018 & 5,014,744 & 48 & 0.38-1.05 $\mu$m & 1 m & 547,807 & 20 & Aerial\\
\bottomrule
\end{tabularx}
\end{table}

Nevertheless these datasets are available and have been shared among researchers for years thanks to the good will of some research groups over the World~\footnote{For example from \url{http://www.ehu.eus/ccwintco/index.php?title=Hyperspectral_Remote_Sensing_Scenes}}. Moreover, {IEEE GRSS} is providing the community with the GRSS Data and Algorithm Standard Evaluation (DASE) website~\footnote{IEEE GRSS DASE website: \url{http://dase.grss-ieee.org/}}. On DASE, researchers can access the data for Indian Pines, Pavia and DFC2018, and submit classification maps which are evaluated on-line. For each dataset, a leader-board allows to compare the state-of-the-art methods as soon as they are tested.

\section{Hyperspectral Image Analysis Issues and Standard Approaches}



This section briefly recalls standard issues and current approaches for hyperspectral data processing. Especially, supervised statistical learning approaches for classification are  detailed since they are obvious reference baselines and inspiration for deep learning-based methods.

\subsection{Pre-processing and normalization}

Working with hyperspectral images often implies pre-processing the data. Besides the aforementioned atmospheric and geometric corrections, band selection and normalization are often also applied. Those normalizations will impact how classifiers are able to separate spectral features in the ways described below.

\subsubsection{Band selection}

Depending on the sensor, some spectral bands might be difficult to process or contain outliers which modify the spectrum dynamics. For example, we often remove bands related to water absorption, bands with a low signal-to-noise ratio and saturated values. Not only this improves the robustness of the classifiers by alleviating the noise present in the data, this also helps fight against the well-known curse of dimensionality that provokes decreasing performances of statistical classification models when the dimensions of the data increases. Band selection can also be used by dropping uninformative bands, e.g. using Principal Component Analysis (PCA)~\cite{farrell_impact_2005} or mutual information~\cite{guo_band_2006}. However, band selection should be done carefully.
Unsupervised dimension reduction can sometimes lead to worse performance than using the raw data since it might remove information that is not useful for compression but was discriminant for classification~\cite{du_band_2003}


\subsubsection{Statistical normalization}



It is a common practice in the machine learning community to normalize the data beforehand to rely on common assumptions for which classifiers are known to behave well, such as zero-mean and unit-variance. Standard strategies often allow to significantly improve processing with statistical approaches. We denote by $X_i$ individual spectra and by $I$ the whole image. These strategies then are:
\begin{itemize}
\item Using the spectral angle, the normalized variant of the spectrum with a unit Euclidean norm:
$X = X / \| X \|$;
The angle between two spectra is a common similarity measure used for classification, notably in the popular Spectral Angle Mapper (SAM) classifier~\cite{yuhas_discrimination_1992}.
\item Normalizing first and second-order moments (so that to obtain a zero mean and unit variance). This can be done for each band independently, which works especially well with classifiers that expect all features to have similar amplitudes, such as Support Vector Machines. However, this squashes the dynamics in the spectral dimension. Alternatively, the normalization can be done globally on the whole image:
$I = \frac{I - m_I}{\sqrt{\sigma_I}}$;
\item Converting the dynamics to $[0,1]$ using  $I = \frac{I - min(I)}{max(I) - min(I)}$. This is helpful mostly for numerical optimization that rarely behave well with very large values and relates more to the implementation than the theoretical standpoint. Once again, this can be applied for each band, at the risk of making relative amplitudes disappear and squashing dynamics, or globally on the whole image.
\end{itemize}

Finally, in order to minimize the influence of outliers and obtain balanced image dynamics, it is also popular to clip on image values over a given threshold: either values over the last decile or outside the range $m_I \pm 2\times \sqrt{\sigma_I}$.


\subsection{Spectral classification}

The most straightforward approach for classifying hyperspectral data is to consider it as a set of 1D spectra. This makes sense given its fine spectral resolution but low spatial resolution. Each pixel corresponds to a spectral signature, that is a discrete signal to which a statistical model can be fit. In the sequel, we restrict our considerations to machine learning approaches with little or not expert processing, but those previously cited.


\subsubsection{Unmixing}

In an ideal case, a pixel in an hyperspectral image corresponds to the reflectance of a material observed over a surface unit. However, the actual spatial resolution of HSI implies that often a pixel corresponds to a surface made of several various materials which produce a spectra mixture. Formally, if we denote by $S_1, \dots, S_n$ the pure spectra (end-members) of the set of materials in the scene, then for a pixel of coordinates $(i,j)$, the locally observed spectrum is a function $F$ of $S_i$:
$$ \phi_{i,j} = F(S_1, \dots, S_n) \simeq \sum_{k = 1}^n \lambda_k S_k~.$$
Under the hypothesis of a plane surface, we may assume $F$ is a mere linear combination where weight coefficients $\lambda_k$ correspond to the proportion of material $k$ in the observed area.


One way to perform hyperspectral data classification is unmixing~\cite{parra_unmixing_1999}, that is finding the individual materials in the observed area by computing their abundance maps. Reference spectra of pure materials are called end-members~\footnote{A mineralogy term form pure minerals, by opposition to most minerals which exist as solid solutions.} and constitute a decomposition basis of mixed spectra. Abundance maps correspond to the proportional contributions of the end-members to each pixel, that are the decomposition coefficients of the mixture. Usually, given the pure spectra $S_k$ and image $\Phi$, it is possible to invert the linear system to obtain the coefficients  $\lambda_k$ for each point, and thus abundance maps. Such approaches rely on linear algebra and numerical methods for solving inverse problems. However, learning-based approaches are also used, for instance clustering methods to find unknown end-members.



\subsubsection{Dimensionality reduction}

Much works are dedicated to reducing the dimension of spectra. Indeed, given the spatial resolution, neighbor intensities are highly correlated so a spectral signature contains a lot of redundant information. The main issue consists in extracting the discriminative information to reduce the set of relevant bands~\cite{le_bris_extraction_2015}.
Among the various pre-processing methods for feature generation which have been used, we can cite feature extraction algorithms such as the well-known principal component analysis (PCA)~\cite{rodarmel_principal_2002} or more recently random feature selection~\cite{damodaran_sparse_2017}.
An other approach consists in computing some indices which integrate physical priors about the band response, such as the Normalized Difference Vegetation Index (NDVI) or the Normalized Difference Water Index (NDWI). This differs from band selection since the resulting features are not reflectances or intensities anymore. Instead, they are a new representation of the data in a space that will be suitable for classification.

Classification is then processed in a standard way, by using common statistical models: decision trees and random forests~\cite{ham-crawford-RF-HSI-TGRS2005}, support vector machines (SVM)~\cite{melgani-bruzzone-HSI-SVM-TGRS2004,gualtieri-cromp-HSI-SVM-SPIE1999}, etc. Approaches such as manifold learning also fit this framework~\cite{chapel_perturbo_2014}. The goal of dimension reduction is to tackle the curse of dimensionality and to simplify the representation space to make the learning stage easier.



\subsection{Spatial-spectral classification}

If a spectral-only approach might work, it is not satisfying since it does not benefit from the spatial structure of hyperspectral images. Indeed, it is likely that neighboring pixels may share some structural relationships (e.g. buildings usually have polygon shapes while vegetation has a fractal-like appearance). Taking the spatial aspect into account during the analysis improves the model robustness and efficiency thanks to these structural dependencies. Three main approaches can be distinguished based on when the spatial aspect is considered in the classification process.


\subsubsection{Spatial regularization}

A popular technique consists to classify individual spectra first, then to regularize the resulting classification with a spatially-structured model such as Markov Random Fields (MRF) or Conditional Random Fields (CRF)~\cite{wu_semi-supervised_2016}. Spatial regularization is then a supervised post-processing.


\subsubsection{Pre-segmentation}

An alternate approach consists in performing spatial regularization as an unsupervised pre-processing. Various methods~\cite{tarabalka_segmentation_2010,fauvel_advances_2013,aptoula_vector_2016} propose
 to first segment the hyperspectral image, then to aggregate spectrum-wise features for each segmented region in order to enforce local consistence. Hierarchical segmentations, e.g. tree-like segmentations, are often used to derive local morphological features, such as in the popular morphological attribute profile approach~\cite{dallamura_extended_2010}.
 
 
\subsubsection{Joint learning}

The last approach learns simultaneously spatial and spectral features by using specific kernels. It takes roots in the inspiring works of \cite{plaza_spatial/spectral_2002} to compute end-members by benefiting from the correlation between spatially close pixels and \cite{dellacqua_exploiting_2004} by exploiting a mixture of spatial and spectral classifiers. More recent approaches focus on statistical models able to learn directly over local neighborhoods (fixed or adaptive) how to extract combined spectral and spatial features. In particular, \cite{camps-valls_composite_2006} introduced the possibility to design spatial-spectral kernels for SVMs able to handle hyperspectral data. This technique will then be largely adopted in later works~\cite{tarabalka_spectralspatial_2009,fauvel_spatial-spectral_2012,cui_scalable_2017}. With a similar objective, \cite{tuia_multiclass_2015} proposes a methood to choose automatically the filters which lead to the most efficient features for hyperspectral data classification from a random-filter bank.

 
The main limitation of traditional shallow learning methods stems from the feature engineering required to improve the classifier's performances. Indeed, spectra from different classes have to be separated in the feature space which can be challenging to achieve. In comparison, deep learning is focused on representation learning, i.e. automatically designing a feature space that is tailored to the objective task. This reduces significantly the need for feature engineering and hopefully should improve performances since both representation and classification will be jointly optimized.
 
\section{Deep Learning for Hyperspectral Data}

Recent works use deep learning techniques for classifying hyperspectral images. The review which follows is organized so as to identify the main families of methods.

\subsection{Pre-processing and normalization}

Pre-processing and normalization processes used for deep learning are similar to the ones used for standard machine learning. However, it is worth noting that most works do not use band selection or saturated spectrum removal but on the contrary rely on the robustness of neural networks.

For unmixing, a now standard, unsupervised  approach~\cite{Guo2015HyperspectralIU,ozkan17endnet} consists in a network with two stacked auto-encoders: the first one is used for denoising while the second one does the actual unmixing by enforcing a sparsity constraint.


\subsection{Spectral classification}

\subsubsection{Supervised learning} The most straightforward evolution from shallow machine learning to deep learning is using a deep fully-connected network instead of a standard classifier (SVM or Random Forest). The principle remains the same, but the network may thematically model the task in a finer way and with a better discrimination capacity. This has been implemented since the 2000s~\cite{goel_classification_2003,ratle_semisupervised_2010} with small networks, and brought up to date recently by~\cite{hu_deep_2015} with unidimensional CNN which learn a filter collection to be applied on individual spectra. However, processing sequential data can also be done using Recurrent Neural Networks (RNN). RNN use memory to retrieve past information and are often used to process time series. \cite{mou_deep_2017} suggests to use these RNN to classify hyperspectral data by assuming that these models can efficiently model both long-range and short-range dependencies in the spectral domain. A similar approach using RNN treating hyperspectral pixels as a sequence of reflectances was introduced in~\cite{wu_semi-supervised_2018}, including a pseudo-labeling scheme for semi-supervised learning. Finally, an approach using both the recurrent and convolutional aspects has been proposed by~\cite{wu_convolutional_2017}, in which the filtered features are finally processed by recurrent layers.


\subsubsection{Unsupervised learning} One of the most important benefits of deep learning for processing hyperspectral data is the introduction of auto-encoders. Indeed, the band selection problem and more generally dimensionality reduction can be considered as a data compression issue. Within this perspective, auto-encoders allow to learn a smart compression with minimal information loss, for example more efficient than a standard PCA. Thus~\cite{xing_stacked_2015} and later~\cite{fu_semi-supervised_2016} proposed dimension reduction through cascade of auto-encoders for denoising followed by classification with a simple perceptron. 


\subsection{Spatial-spectral approaches}


\begin{figure}[!tbp]
\begin{center}
\includegraphics[width=0.48\textwidth]{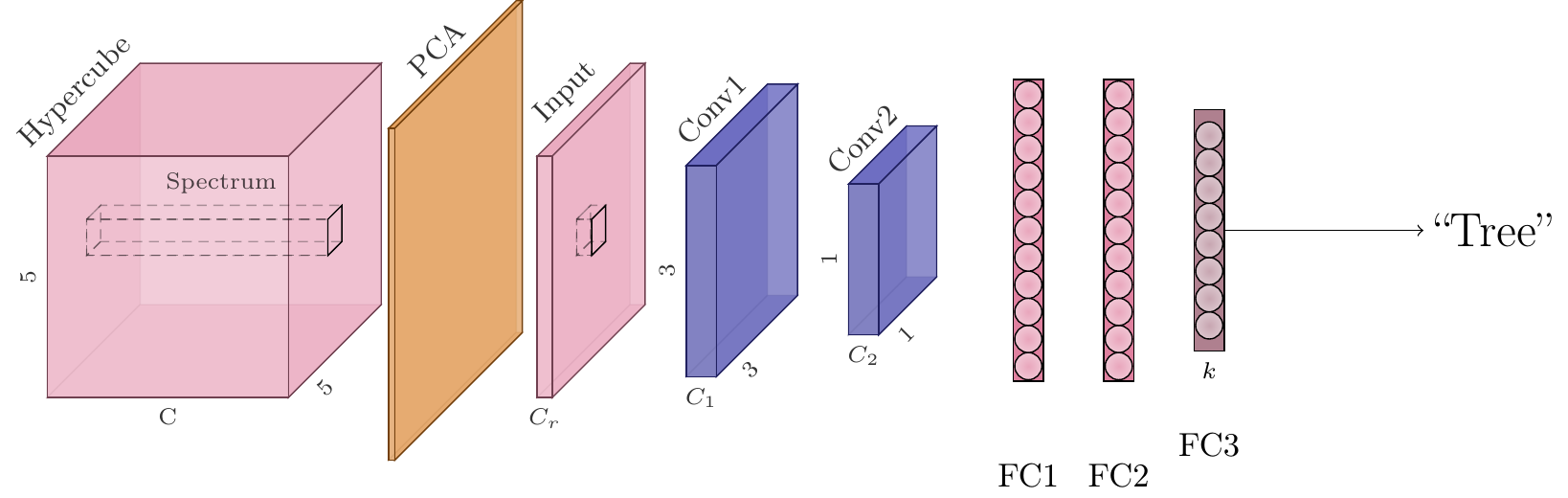}
\caption{2D CNN for classification of hyperspectral data proposed in~~\cite{makantasis_deep_2015}.}
\label{fig:makantasis2d}
\end{center}
\end{figure}



\begin{figure*}[!tbp]
\begin{center}
\includegraphics[width=0.9\textwidth]{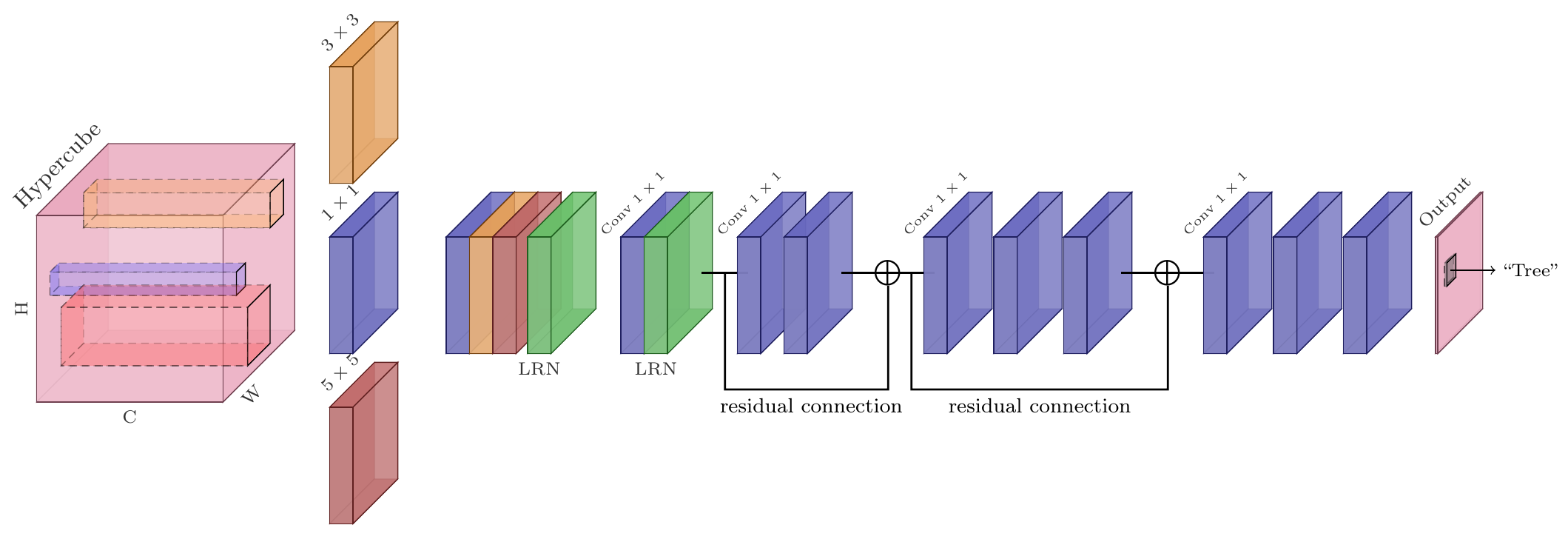}
\caption{2D+1D CNN for classification of hyperspectral data with a residual learning mechanism as proposed in~\cite{lee_kwon-contextualCNN4HSI_TIP2017}.}
\label{fig:lee3d}
\end{center}
\end{figure*}

\subsubsection{1D or 2D Convolutional Neural Networks}
A more recent approach grounds in convolutional networks so popular in multimedia vision: Convolutional Neural Networks (CNN). In computer vision, most CNN are designed using a first part which is convolutional and performs the feature extraction and representation learning, and a second part which is fully connected and performs the classification. However, the number of filters is proportional to the number of input channels, e.g. for a first convolutional layer with a kernel $5\times5$ and $n$ output channels, there will be $5\times5\times{}n\times3$ for an RGB image (3 channels) but $5\times5\times{}n\times100$ for common hyperspectral images (with 100 bands). Therefore, a popular approach to transpose deep convolutional networks to hyperspectral imaging consists in reducing the spectral dimension in order to close the gap between hyperspectral and RGB images.

\paragraph{Supervised learning} Thus, several works proposed a CNN for hyperspectral data classification. \cite{makantasis_deep_2015} uses a PCA to project the hyperspectral data into a 3-channel tensor to then perform the classification using a standard 2D CNN architecture, as shown in~\cref{fig:makantasis2d}. The architecture alternates convolutions and dimension reduction (either by PCA or by sampling) followed by a multi-layer perceptron for the final classification step, as shown in~\cref{fig:makantasis2d}, with 2D convolutions in red and 1D fully connected layers in blue. As an alternative, \cite{slavkovikj_hyperspectral_2015} flattens the spatial dimensions to produce a 2D image with a different shape, instead of an hypercube, and then applies a traditional 2D CNN on the resulting image. One drawback of these methods is that they try to make hyperspectral images similar to RGB ones, i.e. to force hyperspectral data into the multimedia computer vision framework. However, the specific properties of hyperspectral imaging might be wasted when doing so, especially when using unsupervised and uncontrolled dimension reduction.

Trying to go around this problem, \cite{zhao_spectral-spatial_2016,yue_spectral-spatial_2015} propose another approach. Instead of dealing with the hypercube as a whole, they introduce two CNN: a 1D CNN that extracts a spectral feature along the radiometric dimension, and a 2D CNN that learns spatial features as previously described. The features from the two models are then concatenated and fed to a classifier to perform a spatial-spectral classification as usual.

\paragraph{Unsupervised learning} It is worth noting that this kind of models has also been extended for the semi-supervised and unsupervised setups. \cite{zhao_combining_2015} introduced 2D multi-scale convolutional auto-encoders for simultaneous embedding learning of hyperspectral data including both spatial and spectral information and classification, while \cite{romero_unsupervised_2016} proposed the use of CNN for unsupervised feature extraction: it performs reduction dimension of spectra by using also the spatial information using a regularizing sparse constraint.

Similarly to standard approaches, a first approach to spatial-spectral classification consists in representing a pixel by a descriptor with two parts\,:
\begin{itemize}
\item a spectral feature computed on the radiometric spectrum;
\item a spatial feature computed on the local neighborhood of the considered pixel.
\end{itemize}

A standard descriptor is a vector obtained from the concatenation of the complete spectrum and the spatial feature obtained by PCA on a $w\times{h}$-sized neighborhood around the pixel in all bands, from which the $K$ principal components are kept (usually $w = h \simeq 8$
and $K = 3$).
This descriptor is then processed by a deep classifier, often unsupervised such as Deep Belief Networks in~\cite{li_classification_2014,chen_spectral-spatial_2015}, Restricted Boltzmann Machines in~\cite{lin_spectral-spatial_2013,midhun_deep_2014}, or cascades of auto-encoders in~\cite{chen_deep_2014,ma_spectral-spatial_2016,tao_unsupervised_2015,wang_spectralspatial_2017}.

Overall, while the previous approaches introduced deep learning into the hyperspectral imaging framework, they did not fully leveraged the representation learning ability of end-to-end deep networks. Indeed, \emph{ad hoc} processing of the hypercube, either by splitting the dimensions or by unsupervised dimension reduction, could be replaced by learned counterparts.


\subsubsection{2D+1D CNN} 

\paragraph{Supervised learning} Indeed, the importance of spectral information leads to new approaches dealing globally with the hyperspectral cube. The idea is to process directly the hypercube using an end-to-end deep learning process. To answer the problem of the spectral dimension reduction, several works tried to design CNN alternating spatial convolutions with spectral ones to regularly reduce the size of the feature maps.

In particular~\cite{ben_hamida_deep_2016} introduces a CNN which consider both the spatial and spectral neighborhoods of a given pixel, i.e. that takes a 3D patch as an input. The first layers reduce the spectral dimension using a $1\times1\times{}n$ kernel, then the spatial ones with a $k\times{}k\times1$ kernel, and so on. Eventually, two fully-connected layers perform the final classification step. This allows them to compute feature maps where both spectral and spatial representations are learned in an alternate way.

In a similar idea, \cite{luo_hsi-cnn_2018} suggests an alternative approach that performs spatial-spectral convolutions in the first layer to perform spectral dimension reduction, similarly to what could be expected from PCA, albeit supervised and including spatial knowledge. Deeper layers form a traditional 2D CNN that performs as usual.

\paragraph{Unsupervised learning.} On the unsupervised side, \cite{mou_unsupervised_2018} introduce a 2D CNN with a residual learning paradigm~\cite{he_deep_2016} that is able to learn an efficient low-dimension representation of the hyperspectral pixels and their neighborhood.

Finally, \cite{lee_kwon-contextualCNN4HSI_TIP2017} proposes a Fully Convolutional Network (FCN) which handles $N$ bands. The first layer extracts a multi-scale spatial-spectral feature using a module inspired by the Inception architecture~\cite{szegedy_going_2015}. The model applies in the raw data several convolutions with an increasing kernel size in the spatial dimensions, i.e. $1\times1\times{N}$, $3\times3\times{N}$ and $3\times3\times{N}$
where $N$ is the number of bands. This reduces the spectral dimension and also performs a multi-scale filtering of the data. These resulting activation maps are then fed to a succession of non-linearities and 1D convolutions that project the feature maps in the final classification space. This architecture is illustrated in~\cref{fig:lee3d}, with 1D convolution in pink, 2D convolutions in blue and pooling in orange. It is worth noting that thanks to its Fully Convolutional architecture, this network generates predictions for all pixels in the input patch, and not only the central one. This means that it is more efficient at inference time when sliding over the whole image.



\subsubsection{3D CNN} 

If spatial-spectral methods already reach particularly satisfying results, especially with respect to the spatial regularity, they require a high level of engineering in their design which is not fully compatible with the ``data-to-output'' motto that defines deep learning. A promising approach~\cite{li_spectralspatial_2017} handles directly the hyperspectral cube with 3D CNN which work simultaneously on the three dimensions using 3D convolutions. This conceptually simpler approach slightly improves the classification performances with respect to 2D+1D models. Many architectures have been proposed to handle 3D convolutional neural networks for hyperspectral data, mostly to investigate well-known techniques from deep learning for computer vision, such as multi-scale feature extraction~\cite{he_multi_2017} and semi-supervision~\cite{bing_semi_2017}. Instead of producing 2D feature maps, these 3D CNN produce 3D feature cubes that are well-suited for pattern recognition in a volume and seem at least theoretically more relevant for hyperspectral image classification. \cite{chen_deep_2016} especially showed that 3D CNN for classification of hyperspectral images obtained higher performances than their 2D counterparts.

Indeed, with respect to spectral or 2D+1D CNN, 3D CNN combine  those two pattern recognition strategies into one filter, requiring less parameters and layers. They can learn to recognize more complex 3D patterns of reflectances: co-located spectral signatures, various differences of absorption between bands, etc. However, all directions are not equivalent in the hyperspectral data cube, so these patterns cannot be processed as mere volumetric data. As for 2D+1D CNN, it implies extra care when designing the network like using anisotropic filters.

\subsection{Comparative Study of Deep Learning Architectures for Hyperspectral Data}
\subsubsection{Deep Learning for Hyperspectral Toolbox}

\begin{figure*}[!tbp]
\begin{center}
\includegraphics[width=0.95\textwidth]{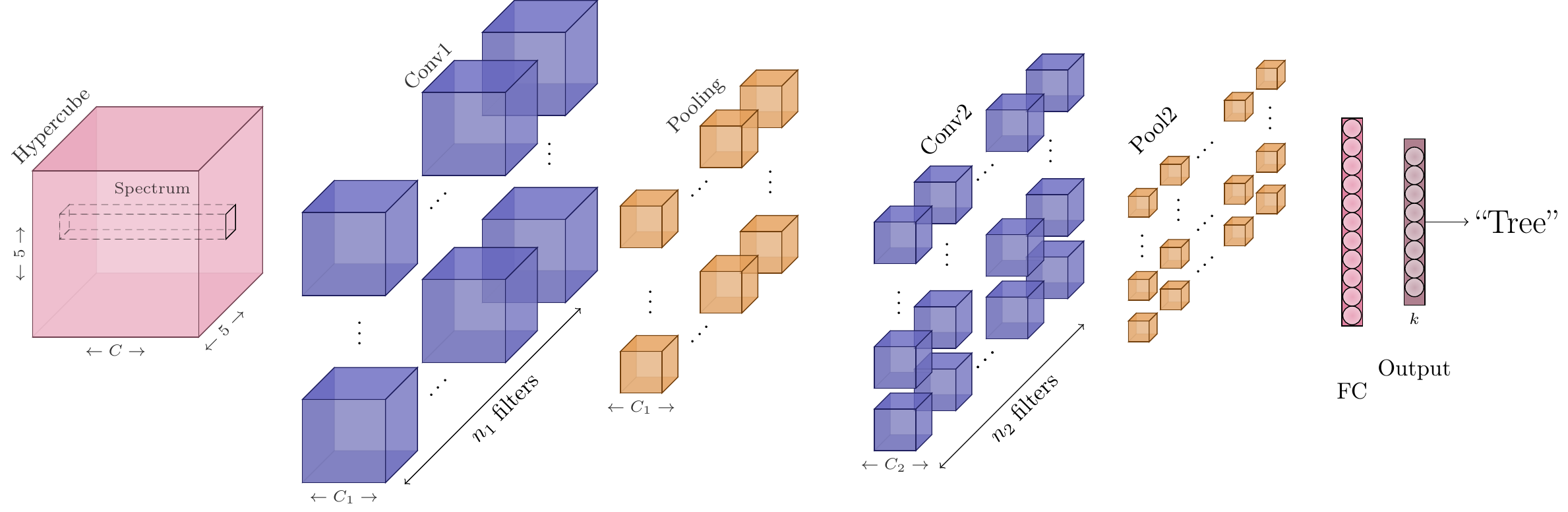}
\caption{3D CNN for classification of hyperspectral data available in the toolbox. It reproduces the architecture proposed in~\cite{chen_deep_2016} and alternates 3D convolutions and 3D max-pooling layers.}
\label{fig:chen3d}---
\end{center}
\end{figure*}

\begin{figure}[!tbp]
\begin{center}
\includegraphics[width=0.49\textwidth]{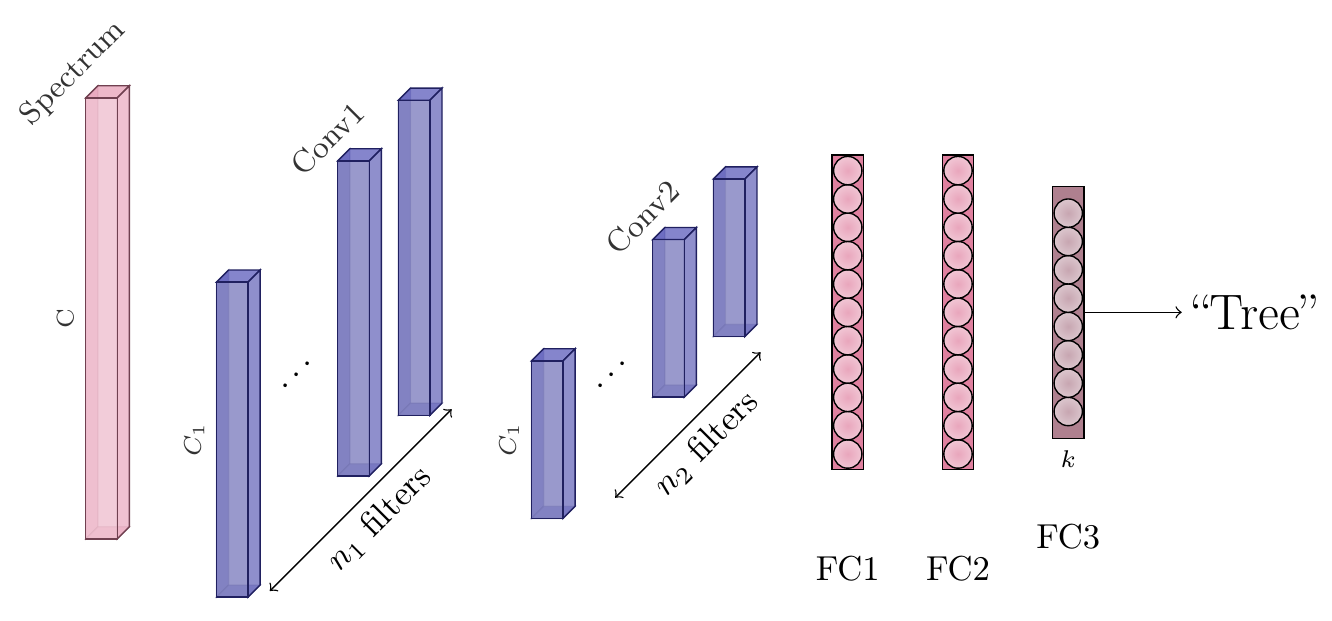}
\caption{1D CNN for spectral classification of hyperspectral data available in the toolbox.}
\label{fig:dnn1d}
\end{center}
\end{figure}

Despite the abundant literature on data driven hyperspectral image processing, there are only few available tools that can be used to compare the state of the art methods in a common framework. To this end, we provide {DeepHyperX}~\footnote{\url{https://github.com/nshaud/DeepHyperX}}, a deep learning toolbox based on the PyTorch framework that can be used to benchmark neural networks on various public hyperspectral datasets. It includes many supervised approaches, ranging from linear SVM to state of the art 3D CNN. Indeed, many models from the literature are implemented in the toolbox, including 1D CNN for spectral classification (cf. Fig.~\ref{fig:dnn1d}, with 1D convolutions in pink, pooling in orange and fully connected layers in blue) and the most recent 2D and 3D CNN as suggested in~\cite{ben_hamida_deep_2016,chen_deep_2016,lee_kwon-contextualCNN4HSI_TIP2017,li_spectralspatial_2017} (cf. Fig.~\ref{fig:chen3d} for a 3D convolutional network architecture example, with 3D convolutions in green, 3D pooling in orange and 1D fully connected layers in blue). The models can be trained and evaluated on several datasets from the literature, e.g. Pavia Center/University, Indian Pines, Kennedy Space Center or DFC2018, described in Sec.~\ref{sec:datasets}. Many hyper-parameters can be tuned to assess the impact of spatial features, the size of the training set or the optimization strategy. This toolbox allows us to evaluate how the methods from the state of the art compare in different use cases.


Technically, this toolbox is written in Python and consists in an interface around the PyTorch and scikit-learn libraries. Deep networks are implemented using PyTorch that can leverage both CPU and GPU based on the user needs, while we rely on scikit-learn for the SVM implementations. Several public classification datasets are pre-configured for easy investigation of deep models. The modular architecture of the toolbox allow users to easily add new remote sensing datasets and new deep architectures.

\subsubsection{Best practices}


Designing and optimizing deep networks can seem arcane, especially considering the abundant literature on this topic that has been published since 2010. However, a principled approach can drastically ease the process of experimenting deep learning on novel applications. A deep learning experiment can be divided in three broad stages: building a model, optimizing the model and running inferences. We are assuming here that the model will be used for classification using a classification loss function such as the cross-entropy.

Building a model can often be summed up as choosing a model from the literature. Most of the time, it is better to rely on a well-validated architecture from the state of the art than spending time designing one from scratch. Based on the dataset and the application, choosing a model can be done on the following criterion:
\begin{itemize}
	\item If the dataset is an image that presents spatial correlations, a 2D approach will outperform pixel-wise classifiers in most cases. For hyperspectral images, 3D CNN will be able to leverage correlations in all three dimensions of the data.
    \item Bigger models mean more parameters to optimize requiring in turn more training samples.
    \item Large convolutional kernels tend to be slower, especially in 3D. Most implementations are optimized for small 2D kernels.
    \item Fully Convolutional Networks are more efficient since they can predict several pixels at a time. Moreover, the absence of fully connected layers means that they will have a lot less parameters, therefore being easier to train.
    \item Non-saturating activation functions such as ReLU alleviate vanishing gradients and help build deeper networks while being faster to compute than sigmoid or $\operatorname{tanh}$ alternatives.
\end{itemize}

During training, the algorithm of choice for optimizing the network's weights is the backpropagation algorithm~\cite{lecun_backpropagation_1989}. This technique relies on the stochastic gradient descent (SGD) or one of its variants. Many versions have been proposed in the literature, using more or less hyperparameters. The fundamental hyperparameter is the learning rate $\alpha$ which controls the magnitude of the weight update after each iteration. A too high $\alpha$ will cause the loss to diverge or to oscillate around the local minimum without reaching it, while a too small $\alpha$ will be very slow to converge. In practice, it is recommended to train with the highest $\alpha$ that makes not the loss diverge at first, and then slowly decreasing it during the training. For example, our toolbox uses an adaptive policy that divides $\alpha$ by one order of magnitude when the validation error plateaus. An alternative is to use an SGD variant with an adaptive learning rate, such as the popular~\cite{kingma_adam:_2015}.

Dealing with the backpropagation with a large number of weights can be complex. Optimizing deep networks involves working with many parameters. One fundamental best practice that is often overlooked is the virtue of initialization. Weights are randomly initialized at the start of the training, but various strategies have been developed so that the SGD starting point has better convergence properties. The initialization policy from~\cite{he_delving_2015} is especially suited for CNN including ReLU activation. Also, due to their large number of weights, the fully connected layers are often prone to overfitting. Dropout~\cite{srivastava_dropout:_2014} significantly alleviates this phenomenon. Moreover, deeper networks often benefit from a larger batch size and the use of Batch Normalization~\cite{ioffe_batch_2015}, which smooths the loss landscape.

Training sample preparation also includes a few practices that can be applied for better performances. First, shuffling the dataset after each epoch helps avoiding recurring patterns in the SGD and overall makes the optimization smoother. Data augmentation is especially useful to introduce equivariances. For example, in image data, horizontal and vertical symmetries can be applied by flipping the training patches randomly during the training. This increases the diversity of the examples and the robustness of the model. Moreover, many datasets present large class imbalance, where one or a few classes dominate the labels. One simple solution is to weight according to the loss function in order to penalize more the less-occurring classes. The inverse-median frequency class weighting is commonly used  to do so, e.g. in semantic segmentation. This is equivalent to showing more examples from the rarer classes to the model.

It is fundamental to be careful when tuning the optimization hyperparameters. Their choice should be based on a validation set that is not the same as the test set, or the test results will be optimistic. If this is not possible, a cross-validation over several train/test splits helps to assess how robust the hyperparameters are in order to avoid overfitting.

Finally, during inference it is recommended to use the network that reached the best validation score and not necessarily the last epoch weights. This implies saving regular checkpoints during the training.

We tried our best in our toolbox to apply these best practices while letting advanced users use their own parameters where needed.

\subsubsection{Experiments}

In this section, we compare several deep architectures from the literature for hyperspectral image classification in a remote sensing context. To the best of our knowledge, there have been none principled analysis of the various deep convolutional networks introduced in the past. Indeed, works from the literature perform experiments using slightly different setups:
\begin{itemize}
	\item Most papers divide the datasets in a train and test splits by randomly sampling over the whole image. A few papers (e.g.~\cite{mou_deep_2017,kemker_low-shot_2018}) use the standard train/test split from the IEEE GRSS DASE initiative.
    \item Some authors consider only a subset of the classes. This is especially prominent for Indian Pines, where the classes with less than 100 samples are often excluded, e.g. in~\cite{hu_deep_2015,lee_kwon-contextualCNN4HSI_TIP2017}.
    \item Even when the train/test splits are done the same way, the number of samples in the train set might vary. Some authors use 20\% of all the training set, while others use a fixed amount of samples for each (e.g. 200 samples for each class).
    \item Some authors further divide the training set into a proper training set and a validation set for hyperparameters tuning, while others perform the tuning directly on the test set.
\end{itemize}

In this work, we argue that randomly sampling the training samples over the whole image is not a realistic use case. Moreover, we affirm that it is a poor indication of generalization power. Indeed, neighboring pixels will be highly correlated, which means that the test set will be very close to the train set. To demonstrate this, we consider a nearest-neighbor baseline using randomly sampled training pixels, and another using co-located training pixels well-separated from the test set.

\begin{figure}
\begin{subfigure}{0.25\textwidth}
	\includegraphics[width=\textwidth]{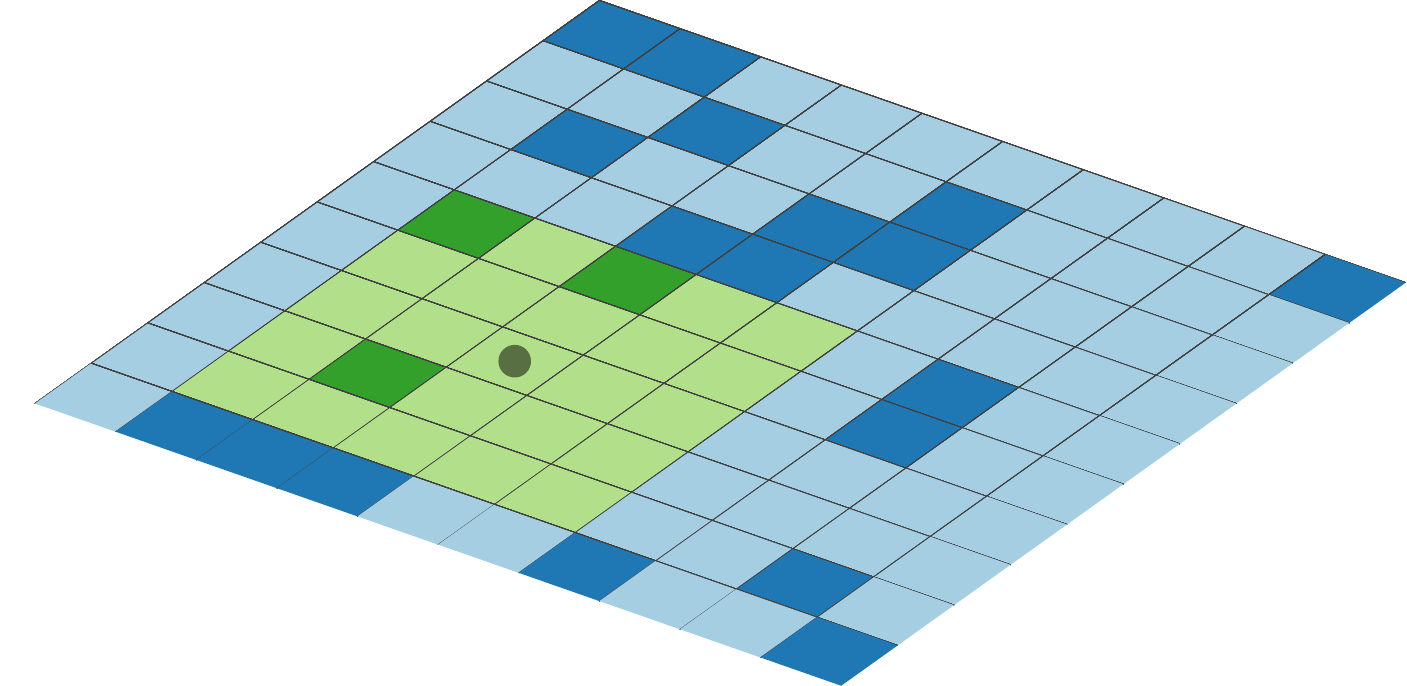}
	\caption*{Random train/test}
\end{subfigure}%
\begin{subfigure}{0.25\textwidth}
	\includegraphics[width=\textwidth]{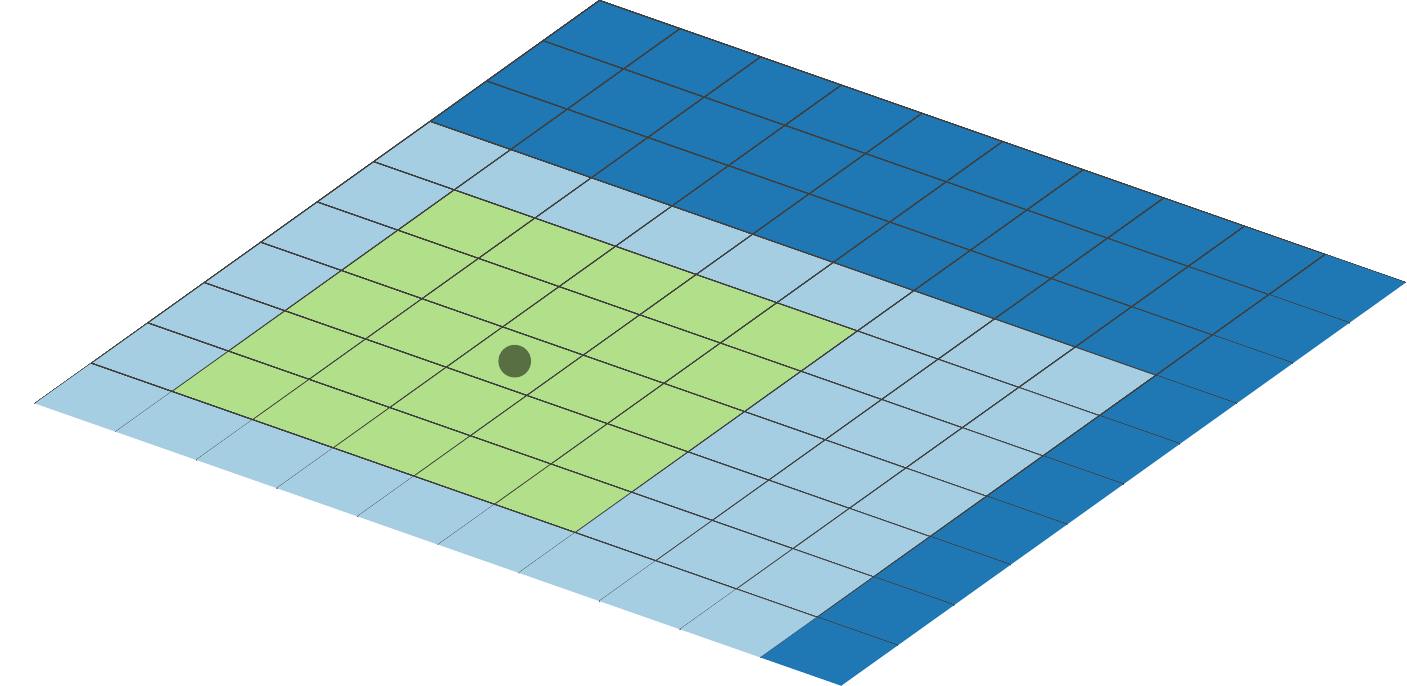}
    \caption*{Disjoint train/test}
\end{subfigure}
\caption{\colorbox{color0}{Train}, \colorbox{color1}{Test}, \colorbox{color2}{Receptive field}, \colorbox{color3}{Test pixel in receptive field}, \textbullet: center pixel.\\The 2D receptive field of a CNN can involuntarily include samples from the test set, making the network overfit and biasing the evaluation.}
\label{fig:receptive_field}
\end{figure}

Also, in the case of 2D and 3D approaches, especially Convolutional Neural Networks, the receptive field of the network might involuntarily includes test samples in the training set. Indeed, the first convolutional layer also sees the neighbors of the central pixel which might be in the test set if the sampling has not been carefully checked. An example of this is illustrated in~\cref{fig:receptive_field}.

Following the standards from the machine learning for remote sensing community, we perform our experiments by using well-defined train/test splits where the samples are extracted from significantly disjoint parts of the image. In the case of 3D CNN, it ensures that no pixel from the test set will be surreptitiously introduced in the train set. To do so, we use the train/test splits for Indian Pines, Pavia University and the DFC2018 as defined by the IEEE GRSS on the DASE benchmarking website. The ground truth is divided based on the connected components instead of the pixels, which allows the evaluation to actually measure how the model generalizes to new geo-entities. Hyperparameters are tuned using 5\% of the train set as a separated validation set.

We use our toolbox to compare various reimplementations of the state of the art. Models have been implemented as close as possible to the original papers~\footnote{It is worth noting that despite the authors' best efforts, reproducing exactly the results using only the papers and without the original implementations is very difficult. Many hyperparameters and implementation tricks are omitted in the manuscripts, and can only be guessed when trying to reproduce the results. Although this is not specific to hyperspectral data, providing code along with papers has been critical to the success of deep learning in many other areas such as computer vision and natural language processing.} We list below the models compared and the changes we performed, if any:
\begin{itemize}
	\item 1D CNN from~\cite{hu_deep_2015}. As the optimizer is not specified in the original paper, we use the standard SGD with momentum.
    \item 1D RNN~\cite{mou_deep_2017}. The authors experiment both with the standard $tanh$ non-linearity and a parameterized variant. We use the former.
    \item 2D+1D CNN~\cite{ben_hamida_deep_2016}. No modification.
    \item 3D CNN~\cite{li_spectralspatial_2017}. No modification.
\end{itemize}
The 2D+1D and 3D CNN are trained using patches of $5\times5$ pixels. The models are compared with two baselines: an SVM --obtained by grid search-- and a fully-connected 1D neural network with three layers interlaced with Dropout~\cite{srivastava_dropout:_2014} using the ReLU non-linearity~\cite{nair_rectified_2010}. To compensate the class unbalance in the three datasets, we use inverse median frequency for all models. We also use vertical and horizontal symmetries as data augmentation for the 2D and 3D CNN. We report in~\cref{tab:results} the overall accuracy and Cohen's kappa on the three datasets. Experiments are repeated 5 times on Pavia University and Indian Pines, though only once on the DFC2018 given its much larger size.



As could be expected, we obtain results significantly lower than those reported in the literature since we use strongly disjoint training and testing sets. \cref{tab:reproduce} reports the results of our implementation, the original score reported by the authors of each model and the gap between evaluating either on a uniform random or a disjoint separated test set. Except for the RNN from~\cite{mou_deep_2017} for which our reimplementation underperforms, all of our reimplementations closely approach the results reported in the original papers. However, experimenting on a disjoint test set, which is the realistic use case, shows that the  accuracies are actually much lower than expected, as reported in~\cref{tab:results}. It is interesting to see that Indian Pines exhibit a very different behavior compared to Pavia University and DFC2018. Indeed, including spatial information using 2D+1D and 3D CNN in Indian Pines actually decreases our classification accuracies. Our hypothesis stems from the fact that Indian Pines has a very low spatial resolution (20~m/px) compared to the two other datasets. This leads to each pixel being actually a mixture of the ground materials. As this dataset is focused on crop classification, this means that each pixel will be an average of the crop reflectance over 400m$^2$, and bringing information from neighboring pixels does not really improve the discriminative power of the model. On the higher resolution Pavia University and DFC2018, 3D spatial-spectral CNN significantly boosts the model's accuracy, respectively by 3\% and 2\% compared to the 1D CNN, which shows the capacity of 3D CNN to efficiently combine spatial and spectral patterns with 3D filters. The DFC2018 is a very challenging dataset because of its large number of classes with high inter-class similarity. In our experiments, the 1D NN suffered from grave overfitting and performs worse than a linear SVM trained by SGD. This is because the test set from the DFC2018 has a large spatial extent and pixels have very different statistical properties compared to the train set. This overfitting is not a problem in Indian Pines and Pavia University, where the train and test sets are very similar and on the same scene, but will become much more important on real large scenes such as Houston from the DFC2018. Interestingly, the 2D+1D CNN seems to fail to catch the spatial information, maybe because of the low dimensions of the spatial kernels. Higher resolution datasets such as the DFC2018 would probably benefit from larger receptive fields to model long-distance spatial relationships between pixels.

\newcommand\res[2]{#1{\scriptsize$\pm$#2}}
\newcommand\bres[2]{\res{\textbf{#1}}{#2}}
\newcommand\bbres[2]{\res{\textit{#1}}{#2}}

\begin{table*}
\begin{tabularx}{\textwidth}{Y Y Y Y Y Y Y}
\toprule
Method & \multicolumn{2}{c}{Indian Pines} & \multicolumn{2}{c}{Pavia University} & \multicolumn{2}{c}{Data Fusion Contest 2018}\\
& Accuracy & $\kappa$ & Accuracy & $\kappa$ & Accuracy & $\kappa$\\
\midrule
SVM & 81.43 & 0.788 & 69.56 & 0.592 & 42.51 & 0.39\\
1D NN & \bres{83.13}{0.84} & \bres{0.807}{0.009} & \res{76.90}{0.86} & \res{0.711}{0.010} & 41.08 & 0.37\\
1D CNN~\cite{hu_deep_2015} & \bbres{82.99}{0.93} & \bbres{0.806}{0.011} & \res{81.18}{1.96} & \res{0.759}{0.023} & \textit{47.01} & \textit{0.44}\\
RNN~\cite{mou_deep_2017} & \res{79.70}{0.91} & \res{0.769}{0.011} & \res{67.71}{1.25} & \res{0.599}{0.014} & 41.53 & 0.38\\
2D+1D CNN~\cite{ben_hamida_deep_2016} & \res{74.31}{0.73} & \res{0.707}{0.008} & \bbres{83.80}{1.29} & \bbres{0.792}{0.016} & 46.28 & 0.43\\
3D CNN~\cite{li_spectralspatial_2017} & \res{75.47}{0.85} & \res{0.719}{0.010} & \bres{84.32}{0.72} & \bres{0.799}{0.009} & \textbf{49.26} & \textbf{0.46}\\
\bottomrule
\end{tabularx}
\caption{Experimental results of various models from our toolbox on the Indian Pines, Pavia University and DFC 2018 datasets using the DASE train/test split. Best results are in \textbf{bold} and second are in \textit{italics}.}
\label{tab:results}
\end{table*}

\begin{table}
\begin{tabularx}{0.5\textwidth}{Y l l l l}
\toprule
Dataset & \multicolumn{2}{c}{Indian Pines} & \multicolumn{2}{c}{Pavia University}\\
\midrule
Model	& Random & Disjoint & Random & Disjoint\\
\midrule
Nearest-neighbor & 75.63 & 67.27 & 89.99 & 57.77\\
\midrule
1D CNN~\cite{hu_deep_2015} (original) & 90.16 & - & 92.56 & -\\
1D CNN (ours)  & 89.34 & 82.99 & 90.59 & 81.18\\
\midrule
RNN~\cite{mou_deep_2017} (original) & 85.7 & - & - & 80.70\\
RNN (ours)  & 79.70 & 62.23 & - & 67.71\\
\midrule
2D+1D CNN~\cite{ben_hamida_deep_2016} (original) & - & - & 94.6 & -\\
2D+1D CNN (ours)  & - & - & 92.39 & 83.80\\
\midrule
3D CNN~\cite{li_spectralspatial_2017} (original) & 99.07 & - & 99.39 & -\\
3D CNN (ours)  & 96.87 & 75.47 & 96.71 & 84.32\\
\bottomrule
\end{tabularx}
\caption{Experimental results with respect to methodological discrepancies between various implementations and evaluation strategies on the Indian Pines and Pavia University datasets.}
\label{tab:reproduce}
\end{table}

\subsection{Perspectives}

Several potential research axes appear for the future. A first one, which follows the works described previously, is the implementation of fully-3D, end-to-end networks for hyperspectral data. Indeed, they are theoretically able to perform the reduction dimension in both the spatial domain and the spectral domain. However, due to the small spatial resolution, a direct transfer of 2D CNN which mainly rely on the spatial context of a full spectrum is not the most promising alternative. On the contrary, the transfer of 3D fully-convolutional networks could extract meaningful local features and tackle this issue.

However, there is also a need for new, larger and more complex (e.g., with more classes) reference hyperspectral datasets for the classification task. Indeed, the community has developed standard approaches which already reach satisfying results on most of them, so it is not anymore possible to distinguish between the new proposed approaches.


A second promising research field is the study of unsupervised and semi-supervised models to overcome the sparsely available annotations of hyperspectral data. For example, 3D spatial-spectral auto-encoders might learn representations of hyperspectral scenes independently of the acquisition context, which can be used for subsequent tasks such as clustering or classification.


Finally, a last coming research field is data synthesis, which would allow to simulate realistic hyperspectral scenes. In the recent years, generative models such as Gaussian Mixture Models or Generative Adversarial Networks (GAN) have been applied to generate new training samples. Initially, the idea was that generative models could approximate the distribution of the deep features in the embedded space. Then, the models infers new samples that could be used to train larger classifiers. \cite{davari_gmm-based_2018} (using GMM) and \cite{he_generative_2017} (using GAN) are two examples of such works. However, thanks to the theoretical improvements to the GAN framework, deep generative models are now able to synthesize from scratch hyperspectral pixels~\cite{audebert_generative_2018} and even small hyperspectral cubes using 3D CNN~\cite{zhu_generative_2018}. This can of course be used for data augmentation, although the benefit of adding new data estimated from a generative model does not seem huge at this point. Nonetheless, generative models will likely be a great asset to estimate image transformations that usually are costly to implement or require specific expert knowledge, such as atmospheric correction, transfer function estimation between sensors or image denoising.



\section{Conclusion}

Deep learning already proved to be efficient for hyperspectral data. 2D or 3D convolutional networks allow to combine spatial and spectral information intuitively and the first published works show state of the art performances, often without expert knowledge about the physics or the sensor.

Today, the main challenge is the scarce availability of massively annotated datasets. Yet, data volume is the key of the success of statistical approaches. A promising path is the development of unsupervised approaches and data augmentation by synthesis in order to overcome this issue and unlock the full potential of deep learning in hyperspectral imaging.

\section*{Acknowledgment}

The authors would like to thank Xavier Ceamanos and Naoto Yokoya for their helpful and valuable insights on various issues examined in this article. The authors would also like to thank the National Center for Airborne Laser Mapping and the Hyperspectral Image Analysis Laboratory at the University of Houston for acquiring and providing the DFC2018 data used in this study, and the IEEE GRSS Image Analysis and Data Fusion Technical Committee. The authors thank Prof. Paolo Gamba from the Telecommunications and Remote Sensing Laboratory, Pavia university (Italy) for providing the Pavia dataset. The authors thank Prof. David Landgrebe and Larry Biehl from  Purdue University, West Lafayette (IN., USA) for providing the Indian Pines dataset. 

Nicolas Audebert's work is funded by the joint Total-ONERA research project NAOMI.

\ifCLASSOPTIONcaptionsoff
  \newpage
\fi




\bibliographystyle{IEEEtran}
\bibliography{nico-zotero,bls-add}

\appendix

\section*{About the authors}

Dr. Nicolas Audebert (M.Eng 2015 Supélec, M.Sc 2015 Université Paris-Sud, PhD 2018 Univ. Bretagne-Sud) is a research scientist working on computer vision and deep learning for remote sensing. His PhD in Computer Science, prepared at ONERA, the French Aerospace Laboratory and the IRISA research institute, focused on the use of deep neural networks for scene understanding of airborne and satellite optical images and Earth Observation. He was awarded the ISPRS Best Benchmarking Contribution at GEOBIA'16 and 2nd Best Student Paper at JURSE'17 prizes.

Dr. Bertrand Le Saux (M.Eng 1999, M.Sc 1999 both at INP Grenoble, PhD 2003 Univ. Versailles / Inria Rocquencourt) is a research scientist in the Information Processing and Systems Department at ONERA, the French Aerospace Laboratory. His research objective is visual understanding by means of data-driven techniques. Currently, his interests focus on developing machine learning and deep learning methods for remote sensing, (flying) robotics and 3D vision. He currently serves as the chair of the IEEE GRSS Image Analysis and Data Fusion Technical Committee.

Prof. S{\'e}bastien Lef{\`e}vre (M.Sc. 1999 at TU Compi{\`e}gne, Ph.D. 2002 at University of Tours, French HDR degree 2009 at University of Strasbourg) is Full Professor in Computer Science in University Bretagne Sud since 2010, where he conducts his researches within the Institute for Research in Computer Science and Random Systems (IRISA). He is leading the OBELIX team dedicated to image analysis and machine learning for remote sensing and Earth Observation (\url{http://www.irisa.fr/obelix}). He has coauthored more than 140 papers in image analysis and pattern recognition. His current research interests are mainly related to hierarchical image analysis and deep learning applied to remote sensing of environment. He was co-chair of GEOBIA 2016 and is co-chair of JURSE 2019.

\end{document}